\pdfoutput=1
\documentclass[11pt]{article}

\usepackage[]{ACL2023}

\usepackage{times}
\usepackage{latexsym}
\usepackage{paralist}

\usepackage[T1]{fontenc}
\usepackage[utf8]{inputenc}

\usepackage{microtype}

\usepackage{inconsolata}
\usepackage{graphicx}
\usepackage{multirow}
\usepackage{multicol}
\usepackage{amssymb}
\usepackage{tabularx}
\usepackage{tablefootnote}

\title{Zero-Shot Dialogue Relation Extraction \\by Relating Explainable Triggers and Relation Names}

\author{
	Ze-Song Xu\quad Yun-Nung Chen\\
	National Taiwan University, Taipei, Taiwan\\
	\texttt{r10922a07@csie.ntu.edu.tw\quad y.v.chen@ieee.org}
}

\begin{document}
\maketitle
\begin{abstract}
Developing dialogue relation extraction (DRE) systems often requires a large amount of labeled data, which can be costly and time-consuming to annotate. In order to improve scalability and support diverse, unseen relation extraction, this paper proposes a method for leveraging the ability to capture triggers and relate them to previously unseen relation names. Specifically, we introduce a model that enables zero-shot dialogue relation extraction by utilizing trigger-capturing capabilities. Our experiments on a benchmark DialogRE dataset demonstrate that the proposed model achieves significant improvements for both seen and unseen relations. Notably, this is the first attempt at zero-shot dialogue relation extraction using trigger-capturing capabilities, and our results suggest that this approach is effective for inferring previously unseen relation types. Overall, our findings highlight the potential for this method to enhance the scalability and practicality of DRE systems.\footnote{Code: \url{https://github.com/MiuLab/UnseenDRE}.}
\end{abstract}

% \includepdf[pages=-]{reform.pdf}

\section{Introduction}
%Relation extraction (RE) is a natural language processing (NLP) task that aims to extract the semantic relationships between arguments mentioned in various types of text data~\citep{zhang-etal-2017-position,DBLP:journals/corr/abs-2010-04829,DBLP:journals/corr/abs-2102-01373,huguet-cabot-navigli-2021-rebel-relation}. The task is to extract relevant information from the text data and represent it in a structured form, such as a knowledge graph, that can be used for downstream applications. 
%Dialogue relation extraction (DRE) is a specialized area within relation extraction (RE) that involves identifying the semantic relationships between arguments in a conversation or dialogue \cite{yu-etal-2020-dialogue}. Recently, papers on DRE have mainly focused on improving the relation extraction performance with diverse methods. For example, \citet{DBLP:journals/corr/abs-2109-04008} constructed a dialogue graph and applies GCN mechanism in order to better identify the relations. \citet{albalak2022d} and \citet{lin-etal-2022-trend} learned how to identify explicit triggers from the data and then improve the DRE performance. More recently, with large language models, \citet{son-etal-2022-grasp} proposed a prompt-based fine-tuning approach to improve the performance of relation extraction on the DialogRE dataset. 

Relation extraction (RE) is a key natural language processing (NLP) task that identifies the semantic relationships between arguments in various types of text data. It involves extracting relevant information and representing it in a structured form for downstream applications~\cite{zhang-etal-2017-position,DBLP:journals/corr/abs-2010-04829,DBLP:journals/corr/abs-2102-01373,huguet-cabot-navigli-2021-rebel-relation}. 
Dialogue relation extraction (DRE) is a specialized area of RE that focuses on identifying semantic relationships between arguments in conversations. Recent DRE research has used diverse methods to improve relation extraction performance, including constructing dialogue graphs \citep{lee2021graph}, identifying explicit triggers \citep{albalak2022d,lin-etal-2022-trend}, and using prompt-based fine-tuning approaches \citep{son-etal-2022-grasp}.
%Dialogue relation extraction (DRE) is a specialized area within RE that involves identifying the semantic relationships between arguments in a conversation or dialogue \cite{yu-etal-2020-dialogue}. Recent research on DRE has mainly focused on improving relation extraction performance with diverse methods, such as constructing a dialogue graph \citep{lee2021graph}, identifying explicit triggers \citep{albalak2022d,lin-etal-2022-trend}, and using prompt-based fine-tuning approaches \citep{son-etal-2022-grasp}.

\begin{table*}
\centering
\begin{tabular}{ll}
\hline
\textbf{DialogRE Relation} & \textbf{Similar DialogRE Relation} \\
\hline
\textsf{per:positive\_impression} & \textsf{per:negative\_impression} \\
\textsf{per:boss} & \textsf{per:subordinate} \\
\textsf{per:children} & \textsf{per:parents} \\
\textsf{gpe:residents\_of\_place} & \textsf{per:place\_of\_residence} \\
\textsf{per:place\_of\_birth} & \textsf{gpe:births\_in\_place} \\
\textsf{org:students} & \textsf{per:schools\_attended} \\
\textsf{per:visited\_place} & \textsf{gpe:visitors\_of\_place} \\
\textsf{per:employee\_or\_member\_of} & \textsf{org:employees\_or\_members} \\
\hline
\end{tabular}
\caption{\label{tab:similar-rel}
\vspace{-1mm}
Similar relation examples in DialogRE.}
\vspace{-3mm}
\end{table*}

\iffalse
\begin{table}
\small
\centering
\begin{tabular}{lll}
\hline
\textbf{DialogRE Relation} & \textbf{DialogRE Relation} \\
\hline
\verb|per:positive_impression| & \verb|per:negative_impression| \\
\verb|per:boss| & \verb|per:subordinate| \\
\verb|per:children| & \verb|per:parents| \\
\verb|gpe:residents_of_place| & \verb|per:place_of_residence| \\
\verb|per:place_of_birth| & \verb|gpe:births_in_place| \\
\verb|per:employee_or_member_of| & \verb|org:employees_or_members| \\
\verb|org:students| & \verb|per:schools_attended| \\
\verb|...| & \verb|...| \\
\hline
\end{tabular}
\caption{\label{tab:similar-rel}
Similar Relation on DialogRE Dataset}
\end{table}
\fi

%The RE tasks require a large amount of labeled data for supervised training, which often involves a significant amount of time for manual annotation. Moreover, models trained on limited data can only predict the relations that they have been trained on and cannot identify similar but unseen relations. Therefore, methods that require only a few labeled examples or no labeled examples at all have gained attention. One example is the prompt-based fine-tuning method \citep{DBLP:journals/corr/abs-2001-07676,DBLP:journals/corr/abs-1912-10165}. Recently, \citet{DBLP:journals/corr/abs-2109-03659} transformed the task of relation extraction into an entailment task and improved zero-shot performance. However, this idea has not yet been adopted on DRE scenarios, because it is not trivial to convert a long conversation into an NLI format.

Supervised training for RE tasks can be time-consuming and expensive due to the requirement for a large amount of labeled data. Models trained on limited data can only predict the relations they have been trained on, making it challenging to identify similar but unseen relations. Hence, recent research has explored methods that require only a few labeled examples or no labeled examples at all, such as prompt-based fine-tuning \citep{DBLP:journals/corr/abs-2001-07676,DBLP:journals/corr/abs-1912-10165}. Additionally, \citet{DBLP:journals/corr/abs-2109-03659} improved zero-shot performance by transforming the RE task into an entailment task. However, this approach has not yet been applied to DRE due to the challenge of converting long conversations into NLI format.

%In our work, we notice that different relations may be dependent as listed in Table~\ref{tab:similar-rel}, such as the \emph{parent-child} relationship. These relations are not independent, but all prior work treated all relations independently and modeled different labels in a multi-class scenario, making the models impossible to handle any unseen relations even when relevant to previously seen relations. Therefore, this paper focuses on enabling zero-shot relation prediction. Specifically, if we encounter an unseen relation during testing, but have previously seen a similar relation, we can relate them through explicitly mentioned trigger words (``children'' $\rightarrow$ ``mom'' $\rightarrow$ ``parents''). 

In this work, we observe that different relations may be dependent on each other, such as the \emph{parent-child} relationship listed in Table~\ref{tab:similar-rel}. Prior work has treated all relations independently and modeled different labels in a multi-class scenario, making it impossible for models to handle unseen relations even if they are relevant to previously seen relations. Therefore, this paper focuses on enabling zero-shot relation prediction. Specifically, if we encounter an unseen relation during testing but have previously seen a similar relation, we can relate them through explicitly mentioned trigger words, such as \textsf{per:children} (seen relation) $\rightarrow$ ``mom'' (trigger) $\rightarrow$ \textsf{per:parents} (unseen relation).

To achieve this, we need to identify the key information of the relation as a tool for relation reasoning during inference. We adopt the approach proposed in \citet{lin-etal-2022-trend}, which achieves remarkable results in DRE by capturing explainable keywords in a dialogue for guiding relation extraction. By leveraging such trigger-capturing capabilities, our proposed model can better deduce unseen relations from known relations and associated triggers. Therefore, the proposed DRE model is more practical, as it can generalize to unseen relations.

%To achieve this, we need to identify the key information of the relation as a tool for relation reasoning during inference. We adopted the approach proposed in \citet{lin-etal-2022-trend}, which achieved remarkable results in dialogue relation extraction. \citet{lin-etal-2022-trend} proposed to capture explainable keywords in a dialogue for guiding relation extraction. By leveraging such trigger-capturing capability, our proposed model can better deduce unseen relations from known relations and associated triggers. That is, the capability of capturing keywords can generalize to unseen relations, making our proposed DRE model more practical.

\section{Proposed Approach}

\begin{figure}[t]
  \centering
  \includegraphics[width=\linewidth]{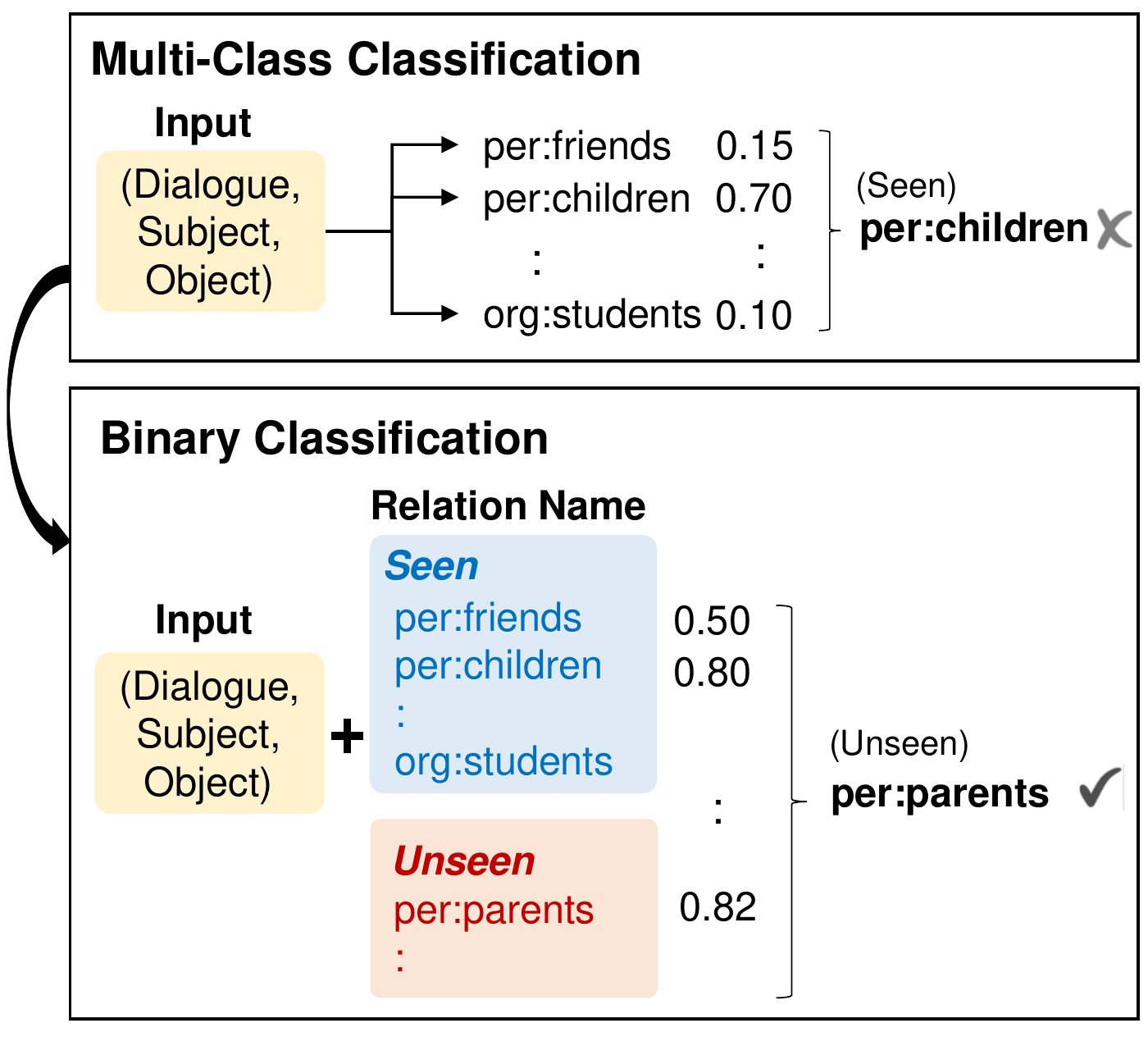}
  \vspace{-3mm}
  \caption{The illustration of our proposed zero-shot relation extraction model.}
  \label{fig:reformulate}
  \vspace{-3mm}
\end{figure}

Prior work on classical DRE has treated it as a multi-class classification problem, which makes it challenging to scale to unseen relation scenarios. To enable a zero-shot setting, we reformulate the multi-class classification task into multiple binary classification tasks by adding each relation name as input, as illustrated in Figure \ref{fig:reformulate}. The binary classification task predicts whether the subject and object in the dialogue belong to the given relation. This approach is equivalent to predicting whether a set of subject-object relations is established, which can estimate any relations based only on their names (or natural language descriptions).

%For classical dialogue relation extraction, all prior work treated it as a multi-class classification problem, making it difficult to scale to unseen relation scenarios. In order to enable the zero-shot setting, we reformulate the multi-class classification task into multiple binary classification tasks by adding each relation name as input as illustrated in Figure \ref{fig:reformulate}. The binary classification is to predict whether the subject and object in the dialogue belong to the given relation. This approach is equivalent to predicting whether a set of subject-object relations is established, which can estimate any relations only based on their names (or natural language descriptions).

\subsection{Model Architecture}
\begin{figure*}
  \centering
  \includegraphics[width=\linewidth]{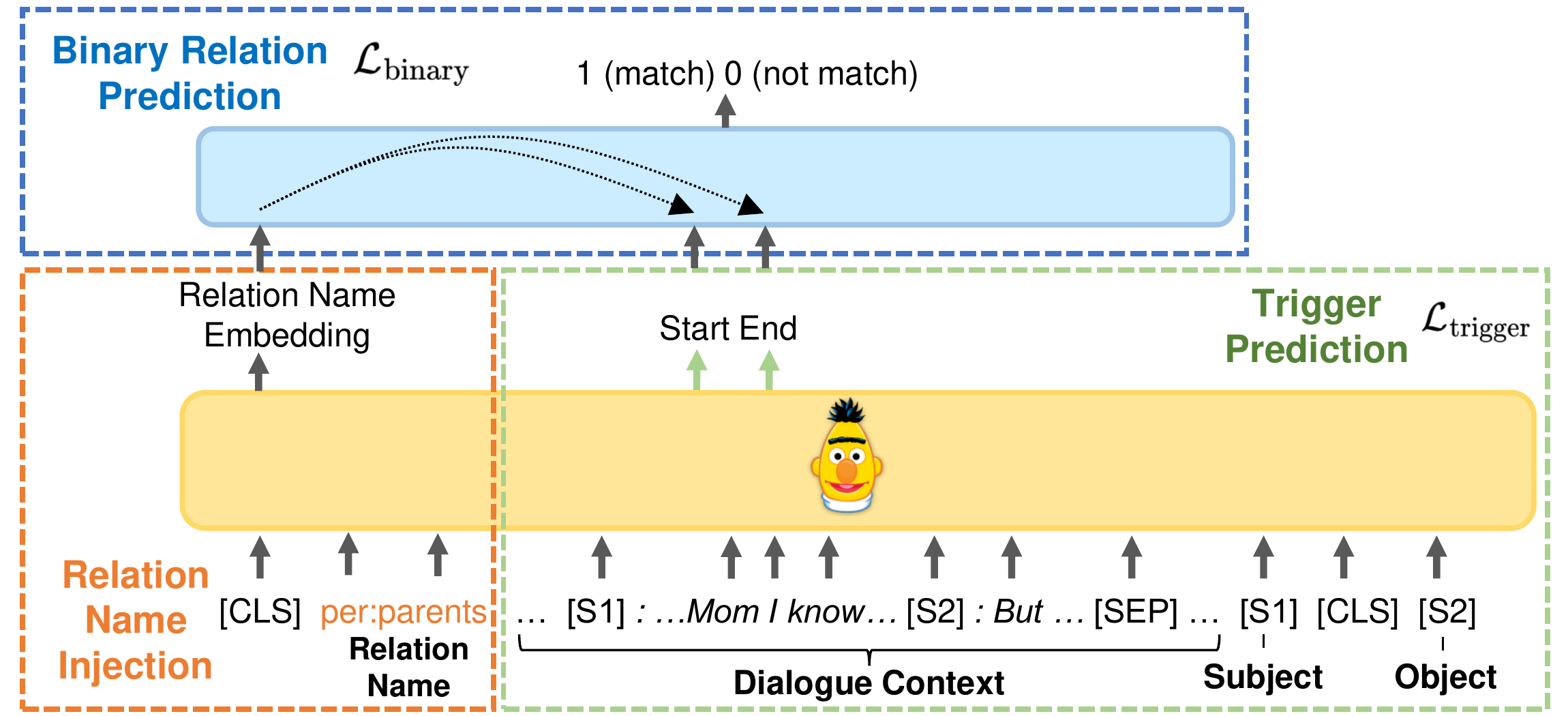}
  \caption{The illustration of our proposed model architecture.}
  \label{fig:model}
  \vspace{-3mm}
\end{figure*}

Our model is illustrated in Figure~\ref{fig:model}, where there are three components in our architecture.

\paragraph{Trigger Prediction}
Inspired by \citet{lin-etal-2022-trend}, we incorporate a trigger predictor into our model, allowing us to employ explicit cues for identifying subject-object relationships within a dialogue. Specifically, we adapt techniques from question-answering models to predict the start and end positions of the trigger span. 
By detecting these triggers, our model not only reasons the potential unseen relations but also enhances the interpretability of the task, making it more practical for real-world applications. 
To identify the keywords associated with \texttt{(Subject, Object, RelationType)} in a dialogue, we formulate the task as an extractive question-answering problem \citep{rajpurkar-etal-2016-squad}. In this setting, the dialogue can be viewed as a document, where the subject-object pair represents the question, and the answer corresponds to the span of keywords that explain the associated relation, i.e., the triggers.

\paragraph{Relation Name Injection}
%Our input format is different from most prior work \citet{DBLP:journals/corr/abs-2109-04008,lin-etal-2022-trend,albalak2022d}, as we also include the relation name after the \texttt{[CLS]} token. This allows the model to have access to natural language descriptions about the given relation, which helps in capturing trigger words more accurately and enabling its zero-shot capability.

In contrast to most prior work \citep{lee2021graph, lin-etal-2022-trend, albalak2022d}, our input format includes the relation name after \texttt{[CLS]}, and we use the \texttt{[CLS]}-associated embeddings as relation name embeddings shown in Figure~\ref{fig:model}.
By doing so, the model has access to \emph{natural language descriptions} of the given relation, which facilitates more accurate capture of trigger words and further enables the zero-shot capability of the proposed model.

\paragraph{Binary Relation Prediction}

In our model, the relation predictor takes as input the learned relation name embedding and a predicted trigger span, as illustrated in the upper part of Figure \ref{fig:model}. To establish the relationship between the relation name and its associated trigger words, we employ a general attention mechanism, where the relation name embedding serves as the query, while the trigger words are encoded by BERT and used as keys and values. The resulting features are then concatenated and fed through a fully connected layer, which generates the final prediction indicating whether the input subject and object have the given relation as expressed in the dialogue.

%The process involves feeding the learned relation name embedding and a predicted trigger span into the relation predictor as shown in the upper part of Figure \ref{fig:model}. To relate the relation name and its trigger words, a general attention mechanism is employed, where the relation name embedding serves as the query, and the trigger words are encoded by BERT and used as the keys and values. These two features are concatenated and passed through a fully connected layer to generate the final prediction indicating if the input subject and object have the given relation shown in the dialogue.
%The input format is "[CLS] Relation Type [SEP] Dialogue [SEP] sub [CLS] obj", and the output is either 0 or 1, depending on whether the subject and object are related.

\subsection{Training}
As depicted in Figure \ref{fig:model}, the input \texttt{(Dialogue, Subject, Oubject, RelationType)} will be initially expanded into a sequence resembling BERT's input format.  The model is trained to perform two tasks. Firstly, it learns the ability to find the trigger span, and secondly, it learns to incorporate the triggers into the relation prediction.

\paragraph{Negative Sampling}

In accordance with \citet{mikolov2013efficient}, we have adopted the negative sampling method in our training process. Specifically, we randomly select some relations from the set of previously observed relations that do not correspond to the given subject-object pair to create negative samples. Notably, the trigger spans of these negative samples remain unchanged.

%Following \citet{mikolov2013efficient}, we employ the negative sampling method. During training, negative samples are created by randomly selecting a relation from the set of seen relations that does not match the subject-object pair. Here the trigger spans of the selected negative samples remain the same.
% In the training data of seen relations, we randomly select relation that do not match the seen relation as negative sample.

\paragraph{Multi-Task Learning}
The trigger prediction task involves identifying the most likely trigger positions, and is treated as a single-label classification problem using cross-entropy loss $\mathcal{L}_{Trigger}$. On the other hand, the relation prediction task employs binary cross-entropy loss $\mathcal{L}_{Binary}$ to compute the prediction loss. To train the model simultaneously on both tasks, we employ multi-task learning. We use a linear combination of the two losses as the objective function. This enables us to train the entire model in an end-to-end fashion. %, resulting in good DRE performance together with the capability of trigger capturing.

%Trigger prediction is a single-label classification problem of predicting the most possible positions through a cross entropy loss $\mathcal{L}_{Trigger}$. In relation prediction, a binary cross entropy loss $\mathcal{L}_{Binary}$ is utilized to calculate the prediction loss. By utilizing multi-task learning, it is possible to simultaneously train these two previously mentioned losses together. A linear combination was used as the learning objective to train the whole model in an end-to-end manner, and the combination weight can be tuned on our development set.

\subsection{Inference}
During inference, our model follows a similar setting to the one used during training. However, we have observed that the model tends to predict the seen relation when the captured trigger words are present in the training data. To prevent the model from overfitting to the seen relations, we replace the trigger span with a general embedding (the embedding of \texttt{[CLS]}), which is assumed to carry the information of the entire sentence. This embedding is used as the input for our relation prediction. By doing so, our model can better generalize to unseen scenarios and can avoid the tendency to predict the seen relation when capturing seen trigger words. This approach enhances the model's ability to handle diverse unseen relations during inference.

%During inference, we follow a similar setting as training. However, it is found that the model tends to predict the seen relation when the captured trigger words are seen. In order to prevent the model from overfitting seen relation, our model replaces the trigger span with a general embedding (embedding of \texttt{[CLS]}), which is assumed to carry whole-sentence information, as the input of our binary relation prediction. By doing so, our model can better generalize to unseen scenarios.

% The inference setting is similar to the training setting, except that the predicted trigger word is replaced with the [CLS] token during inference. This is because using the predicted trigger word during inference may cause the model to be overly focused on the seen relations. By using the [CLS] token, we aim to evaluate the model's performance in a more general setting.

\begin{table*}
\centering
%\begin{minipage}{\textwidth}
\begin{tabular}{|l||c|c|c|c||c|c|}
\hline 
\multirow{2}{*}{\textbf{Model}} & \multicolumn{2}{c|}{\bf Unseen} & \multicolumn{2}{c||}{\bf Seen} & \multicolumn{2}{c|}{\bf Overall\footnotemark[2]{}} \\
 \cline{2-7}
 & \textbf{Top 1} & \textbf{Top 2} & \textbf{Top 1} & \textbf{Top 2} & \textbf{Top 1} & \textbf{Top 2} \\
\hline
Multi-class BERT & ~~0.0 & ~~0.0 &  60.6 & - & 48.5 & -  \\
TUCORE-GCN~\cite{lee2021graph} & ~~0.0 & ~~0.0 & 65.5\footnotemark[1]{} & - & 48.4\footnotemark[1]{} & -\\
TREND~\cite{lin-etal-2022-trend} & ~~0.0 & ~~0.0 & \bf 66.8\footnotemark[1]{} & - & 53.4\footnotemark[1]{} & -  \\
Binary-Reformulated BERT & 24.5 & 28.9 & 57.0 & 45.5 & 50.5 & 42.2 \\
Proposed (with predicted triggers)  & 23.5 & \bf 34.8 & 66.7 & \bf 51.5 & 58.0 & \bf 48.2 \\
Proposed (with relation name embeddings) & \bf 32.5 & \bf 34.8 & 65.6 & 51.0 & \bf 60.0 & 47.8 \\
\hline
Proposed with gold triggers & 35.6 & 40.4 & 70.4 & 53.2 & 63.4 & 50.6  \\
\hline
\end{tabular}
\caption{\label{unseen-performance}
The micro-F1 performance of DialogRE in terms of unseen, seen, and overall settings (\%).}
\vspace{-3mm}
\end{table*}

\section{Experiments}

%\subsection{Setting}
We conducted experiments using the DialogRE dataset, which is widely used as a benchmark in the field. To assess our model's zero-shot capability, we divided the total of 36 relations into 20 seen and 16 unseen types detailed in the Appendix. We only train our model on data related to seen relation types. During training, we set the learning rate to 3e-5 and used a GeForce RTX 2080 Ti. The training process involves 10 epochs without early stopping\footnote{The models with early stopping achieve similar performance.}, and the number of negative samples was 3. To ensure a fair comparison with prior work \cite{lin-etal-2022-trend,yu-etal-2020-dialogue}, we use the same testing set for evaluation.

%In our experiments, we use a benchmark DRE dataset, DialogRE. In order to examine the zero-shot capability, we segment total 36 relations into 20 seen and 16 unseen types and limit our model to only train on the data of seen relation types. The detailed sets are listed in Appendix. Our model is trained with a learning rate of 3e-5 on a GeForce RTX 2080 Ti. The training process performs 10 epochs without an early stop, and the number of negative samples is 3. To fairly compare with the prior work, the testing set is the same as one in the prior work \cite{lin-etal-2022-trend,yu-etal-2020-dialogue}. 

\subsection{Evaluation Metric}

After performing multiple binary classification tasks, our model can rank the relation candidates based on their predicted scores. Typically, the model outputs the relation with the highest score, as done in prior work, and micro-F score is calculated for evaluation. However, since our task is focused on zero-shot performance, we are also interested in whether our model can correctly rank the unseen relations, even if the top-ranked relation is incorrect. To better understand how our model estimates all relation candidates, we evaluate our model not only on the top-ranked relation but also on the top-2 ranked relations in our experiments. This allows us to gain insight into how well our model can rank the correct relations, even if they are not the top-ranked ones.

%We evaluate the performance of our model using micro-F, which is equivalent to the Rank 1 metric used in our experiments. In cases where the model needs to choose between similar relations in unseen scenarios, it tends to prioritize seen relations as the top predicted value. This can cause the model to predict the correct but previously unseen relation as the second-ranked prediction instead of the top-ranked prediction. To observe this behavior, we calculate the micro-F score based on the first two predictions made by the model. This is referred to as the Rank 2 metric used in our experiment.
% The model's ranking performance is evaluated using the Micro-F metric, which is equivalent to the Rank 1 metric used in the experiment. In the scenario where the model needs to make a choice between similar relations in unseen cases, it tends to prioritize seen relations as the top predicted value. As a result, the model may predict the correct relation that has not been seen before as the second-ranked prediction instead of the top-ranked prediction. To observe this behavior, we calculate the micro-F score based on the first two predictions made by the model, which is referred to as the Rank 2 metric used in the experiment.

% As illustrated in  Table~\ref{tab:seen-unseen-rel}, We manually categorized relations as either seen or unseen, with the criterion of separating two similar relations into different categories. For unrelated relations, they were assigned randomly to either category. 

\subsection{Model Setting}
We perform different model settings on BERT-Base for fair comparison.
\begin{compactitem}
    \item \textbf{Multi-class BERT} is a baseline, where BERT-Base  \citep{devlin-etal-2019-bert} is adopted and treated DRE as multi-class classification.
    \item \textbf{TUCORE-GCN} construct a dialogue graph to utilize the graph strucutre for prediction \citep{lee2021graph}.
    \item \textbf{TREND} proposed to capture explicit triggers for better performance \cite{lin-etal-2022-trend}.\footnote{The scores are reported from the prior work for reference, which cannot be directly compared with our scores.}
    \item \textbf{Binary-reformulated BERT} performs  binary classification shown in Figure~\ref{fig:reformulate}, which is a proper baseline for zero-shot settings.
    \item \textbf{Proposed} has three settings in binary relation prediction during inference: 1) based on predicted triggers, 2) based on relation name embddings, 3) based on gold triggers.
    The third is listed as an upper bound for reference.\footnote{Overall performance is estimated based on data size.}
    %taking a weighted average of the seen and unseen data, with a weight of 80\% assigned to the seen data and a weight of 20\% assigned to the unseen data.}
\end{compactitem}

 % the ground truth's embedding is used instead of the predicted trigger embedding during inference.

\subsection{Results}

%\paragraph{Zero-Shot Performance}
Table~\ref{unseen-performance} presents our results. Prior work achieves micro-F scores above 60\% for seen relations but cannot predict unseen relations (0\%) due to their multi-class formulation. The reformulated BERT serves as the baseline for zero-shot settings, achieving 24.9\% and 28.9\% for top 1 and top 2 ranked relations, respectively.

Our proposed method of inputting predicted triggers for relation prediction did not rank correct unseen relations as top 1 (23.5\% vs. 24.5\%). However, the performance of top 2 ranked relations significantly improved (from 28.9\% to 34.8\%), suggesting that trigger prediction is indeed useful. The lower top 1 relations score can be attributed to similar triggers for relevant relations, which easily favor seen relations. An example of incorrect prediction is provided in Table~\ref{tab:qualitative-example}.

Replacing predicted triggers with relation name embeddings, our proposed model achieves the best performance for unseen relations (32.5\% for top 1 and 34.8\% for top 2). This indicates that this setting avoids overfitting to seen relations and allows prediction to better generalize to unseen scenarios.

Moreover, feeding gold triggers into relation extraction during inference yields the best results, indicating the potential for improvement with the proposed trigger mechanism.
In sum, the experiments demonstrate that our proposed model can connect trigger words with relation names and enables zero-shot relation extraction.

%To observe this phenomenon, we used rank2 as the metric and surprisingly found that the predicted trigger word's F1 score was even higher than that of the [CLS] token. This result confirmed our initial observation. Therefore, we replaced the predicted trigger word with the [CLS] token, which contains general information, during inference. This approach significantly improved the rank1 performance.

%\paragraph{[CLS] vs Ground Truth Embedding}
%It is worth noting that using the embedding of the ground truth trigger word can yield better results than using the [CLS] embedding. The above scenario implies that there is still room for improvement in trigger word generation

In terms of performance on seen data, our proposed models outperform the reformulated BERT baseline by a significant margin.
Moreover, our models achieve comparable scores to previous work (66.7\% vs. 66.8\% in top 1 scores), even though we consider more candidates.
These results further validate the effectiveness of our model and its superior generalization capability.

%In terms of performance on seen data, our proposed models achieve significantly better performance than the reformulated BERT baseline by a large margin. Interestingly, the scores of our proposed methods are comparable with ones of previous work (66.7\% vs. 66.8\% in top 1 scores), even though our model considers more relation candidates. The results further demonstrate the effectiveness of our proposed model and its great generalization capability.

After comprehensive analysis, we found that our proposed method incorporating a general context embedding not only leverages the trigger capturing capability but also assists the DRE task indirectly, leading to the best overall performance among all proposed models. The ability to relate trigger keywords to relation names enables the model to generalize better to unseen relations and overcome the limitations of relying on specific trigger words. The results of our experiments demonstrate the effectiveness of our proposed method and its potential for real-world applications.

\subsection{Qualitative Study}
Table~\ref{tab:qualitative-example} showcases an example about the predicted triggers and relations for the DialogRE dataset. As an instance, Sal is the uncle of Speaker 3, so the relation between them should be ``other\_family''. Although the trigger word mechanism accurately captures the crucial keyword ``uncle'', the model still outputs the ``children'' relation from the seen relation category rather than the ``other\_family'' relation from the unseen relation category. This suggests that while capturing significant subject and object information through trigger words, the model tends to prioritize predicting relations from the seen relation category.

\begin{table} [t!]
    \centering
    \small
    \begin{tabularx}{\linewidth}{|X|}
    \hline
    S1: What about Ben? We can't bring a baby to a hospital. \\
    S2: We'll watch him. \\
    S1: I don't think so. \\
    S3: What? I have seven Catholic sisters. I've taken care of  hundreds of kids. Come on, we wanna do it, don't we? \\
    S2: I was looking forward to playing basketball, but I guess that's out the window. \\
    S1: Ok, well, if you do take him out for his walk, you might wanna bring his hat, and there's extra milk in the fridge, and there's extra diapers in the bag. \\
    S3: Hat, milk, got it.\\
    S1: ??? Thro up a thro thro--a thro thro!\\
    S3: Consider it done. \\
    S2: You understood that? \\
    S3: Yeah, my uncle Sal has a really big tongue. \\
    S2: Is he the one with the beautiful wife? \\
    \hline
    (Subject, Object) : (Sal, S3) \\
    Predicted trigger: uncle \\
    Gold trigger: uncle \\
    Predicted relation: per:children \\
    Gold relation: per:other\_family \\
    \hline
    \end{tabularx}
    \caption{\label{tab:qualitative-example}
    An incorrectly-predicted example.}
\end{table}

% {'y': 'Speaker 3', 'x': 'Sal', 'rid': [13], 'r': ['per:other_family'], 't': ['uncle'], 'x_type': 'PER', 'y_type': 'PER'}

% \begin{table}
% \centering
% \begin{tabular}{lc}
% \hline
% \textbf{DialogRE Relation} & \textbf{DialogRE Relation}\\
% \hline
% \verb|TREND-Predict | & \verb|TREND-Predict | \\
% \verb|TREND-Predict | & \verb|TREND-Predict | \\
% \verb|TREND-Predict | & \verb|TREND-Predict | \\
% \hline
% \end{tabular}

% \caption{Example commands for accented characters, to be used in, \emph{e.g.}, Bib\TeX{} entries.}
% \label{tab:accents}
% \end{table}

% \begin{table*}
% \centering
% \begin{tabular}{lll}
% \hline
% \textbf{Model} & \textbf{Rank 1 Micro-F} & \textbf{Rank2 Micro-F}\\
% \hline
% \verb|Baseline-bert| & 0.57 & 0.46 \\
% \verb|TREND-[CLS] token embedding| & 0.66 & 0.51 \\
% \verb|TREND-Predict trigger embedding| & 0.67 & 0.52 \\
% \verb|TREND-Ground Truth trigger embedding| & 0.70 & 0.53 \\
% \hline
% \end{tabular}
% \caption{\label{seen-performance}
% The model performance on Seen Relation}
% \end{table*}

\section{Conclusion}

This paper introduces a novel approach for zero-shot dialogue relation extraction by relating explainable trigger words and relation names. Our proposed method effectively utilizes trigger-capturing capability and demonstrates a significant improvement in inferring unseen relations. The experimental results on benchmark data show that our approach achieves better generalization and practicality, making it a promising solution for real-world applications.

\section*{Acknowledgements}
We thank the reviewers for their insightful comments.
This work was financially supported by the
Young Scholar Fellowship Program by the National
Science and Technology Council (NSTC) in Taiwan,
under Grants 111-2222-E-002-013-MY3 and
111-2628-E-002-016.

% \subsection{References}
% \nocite{Ando2005,augenstein-etal-2016-stance,andrew2007scalable,rasooli-tetrault-2015,goodman-etal-2016-noise,harper-2014-learning}

% The \LaTeX{} and Bib\TeX{} style files provided roughly follow the American Psychological Association format.
% If your own bib file is named \texttt{custom.bib}, then placing the following before any appendices in your \LaTeX{} file will generate the references section for you:
% \begin{quote}
% \begin{verbatim}
% \bibliographystyle{acl_natbib}
% \bibliography{custom}
% \end{verbatim}
% \end{quote}
% You can obtain the complete ACL Anthology as a Bib\TeX{} file from \url{https://aclweb.org/anthology/anthology.bib.gz}.
% To include both the Anthology and your own .bib file, use the following instead of the above.
% \begin{quote}
% \begin{verbatim}
% \bibliographystyle{acl_natbib}
% \end{verbatim}
% \end{quote}
% Please see Section~\ref{sec:bibtex} for information on preparing Bib\TeX{} files.

% \subsection{Appendices}

% Use \verb|\appendix| before any appendix section to switch the section numbering over to letters. See Appendix~\ref{sec:appendix} for an example.

\iffalse
\section*{Acknowledgements}
We thank reviewers for their insightful comments.
This work was financially supported from the
Young Scholar Fellowship Program by Ministry
of Science and Technology (MOST) in Taiwan,
under Grants 111-2628-E-002-016 and 111-2634-
F-002-014.
\fi

\bibliography{anthology,custom}
\bibliographystyle{acl_natbib}

\appendix

%\section{Appendix}
%\label{sec:appendix}

\section{Criteria for Relation Dividing}

We categorized the relations into two sets, namely, seen and unseen, as presented in Table~\ref{tab:seen-unseen-rel}. Our categorization was based on the similarity of relations, where dependent ones are assigned to different categories.
For those not related, we assigned them randomly to either category. This categorization aims to train the model on seen relations to enhance its ability to predict unseen relations during testing.

%As shown in Table~\ref{tab:seen-unseen-rel}, we manually catogorize the relations into two sets: seen and unseen. Our criterion for distinguishing between the two categories was to assign similar relations to different categories. For relations that are not related, we randomly assigned them to either category. This approach aims to enhance the model's ability to predict unseen relations during testing by training it on seen relations.

\begin{table}[t!]
\centering
\small
\begin{tabular}{|ll|}
\hline
\textbf{Seen Relations} & \textbf{Unseen Relations} \\
\hline
\textsf{per:positive\_impression} & \textsf{per:subordinate} \\
\textsf{per:client} & \textsf{gpe:visitors\_of\_place} \\
\textsf{per:origin} & \textsf{per:place\_of\_residence} \\
\textsf{per:works} & \textsf{per:schools\_attended} \\
\textsf{per:place\_of\_work} & \textsf{per:parents} \\
\textsf{per:title} & \textsf{gpe:births\_in\_place} \\
\textsf{per:alternate\_names} & \textsf{org:employees/members} \\
\textsf{per:acquaintance} & \textsf{per:dates} \\
\textsf{per:alumni} & \textsf{per:other\_family} \\
\textsf{per:friends} & \textsf{per:siblings} \\
\textsf{per:girl/boyfriend} & \textsf{per:spouse} \\
\textsf{per:neighbor} & \textsf{per:negative\_impression} \\
\textsf{per:roommate} & \textsf{per:age} \\
\textsf{per:boss} & \textsf{per:date\_of\_birth} \\
\textsf{per:children} & \textsf{per:major} \\
\textsf{gpe:residents\_of\_place} & \textsf{per:pet} \\
\textsf{per:place\_of\_birth} &  \\
\textsf{per:visited\_place} &  \\
\textsf{per:employee/member\_of} &  \\
\textsf{org:students} &  \\
\hline
\end{tabular}
\caption{\label{tab:seen-unseen-rel}
Seen and unseen relations in our experiments.}
\end{table}

\begin{table}[t!]
\centering
\begin{tabular}{|l|c|c|}
\hline 
\multirow{2}{*}{\textbf{Unseen Relation}} & \multicolumn{2}{c|}{\bf Unseen}  \\
 \cline{2-3}
 & \textbf{Predict} & \textbf{CLS} \\
\hline
\textsf{per:siblings} & 26 & 42 \\
\textsf{per:spouse} & 21 & 30 \\
\textsf{per:negative\_impression} & 4 & 11 \\
\textsf{per:parents} & 5 & 9  \\
\textsf{per:dates} & 0 & 4  \\
\textsf{per:major} & 2 & 2 \\
\textsf{per:age} & 1 & 1 \\
\textsf{gpe:births\_in\_place} & 0 & 0  \\
\textsf{org:employees/members} & 0 & 0  \\
\textsf{per:other\_family} & 0 & 0 \\
\textsf{per:date\_of\_birth} & 0 & 0 \\
\textsf{per:pet} & 0 & 0 \\
\textsf{per:subordinate} & 0 & 0 \\
\textsf{gpe:visitors\_of\_place} & 0 & 0 \\
\textsf{per:place\_of\_residence} & 0 & 0 \\
\textsf{per:schools\_attended} & 0 & 0 \\
\hline
\end{tabular}
\caption{\label{unseen-distribution}
The distribution of correct predictions in the predict trigger method and cls trigger method.}
\end{table}

\section{Prediction Distribution Comparison}

We analyze the distribution of correctly predicted top 1 unseen relations for two models, one with predicted triggers and the other with relation name embeddings, and present the results in Table~\ref{unseen-distribution}. We observe that the two methods exhibit a similar pattern of correctly predicted relations, with a concentration on particular unseen relations such as siblings and spouses, among others. However, the proposed method with the relation name embeddings significantly outperforms the one with the predicted triggers method in this aspect.

\end{document}